\documentclass{article}

\usepackage{PRIMEarxiv}

\usepackage[utf8]{inputenc} 
\usepackage[T1]{fontenc}    
\usepackage{hyperref}       
\usepackage{url}            
\usepackage{booktabs}       
\usepackage{amsfonts}       
\usepackage{nicefrac}       
\usepackage{microtype}      
\usepackage{lipsum}
\usepackage{fancyhdr}       
\usepackage{graphicx}       
\usepackage{amsmath}        
\graphicspath{{media/}}     

\DeclareMathOperator*{\argmax}{argmax}

\title{Unstructured Road Segmentation using Hypercolumn based Random Forests of Local experts
}

\author{
  Prassanna Ganesh Ravishankar\\
  Independent Researcher \\
  \texttt{me@prassanna.io} \\
   \AND
   Antonio M. Lopez \\
   Computer Vision Center \\
   Barcelona \\
   \texttt{antonio@cvc.uab.es} \\
   \And
   Gemma M. Sanchez \\
   Computer Vision Center \\
   Barcelona \\
   \texttt{gemma@cvc.uab.es} \\
}

\begin{document}
\maketitle

\begin{abstract}
Monocular vision based road detection methods are mostly based on machine learning methods, relying on classification and feature extraction accuracy, and suffer from appearance, illumination and weather changes. Traditional methods introduce the predictions into conditional random fields or markov random fields models to improve the intermediate predictions based on structure. These methods are optimization based and therefore resource heavy and slow, making it unsuitable for real time applications. We propose a method to detect and segment roads with a random forest classifier of local experts with superpixel based machine-learned features. The random forest takes in machine learnt descriptors from a pre-trained convolutional neural network - VGG-16. The features are also pooled into their respective superpixels, allowing for local structure to be continuous. We compare our algorithm against Nueral Network based methods and Traditional approaches (based on Hand-crafted features), on both Structured Road (CamVid and Kitti) and Unstructured Road Datasets. Finally, we introduce a Road Scene Dataset with 1000 annotated images, and verify that our algorithm works well in non-urban and rural road scenarios. 
\footnote{Prassanna Ganesh Ravishankar implemented the Random Forest module, using the Microsoft's Sherwood library as the starting point. All other code used in this paper were also implemented by Prassanna Ganesh Ravishankar. Antonio M. Lopez was instrumental in designing the experiments, and gave the general trajectory of research. Gemma M. Sanchez, was instrumental in collecting the dataset - which was done by Prassanna Ganesh Ravishankar's Camera mounted on her baby's pram.}
\end{abstract}


\section{Introduction}
Road detection is an important step in a pipeline for an autonomously driven vehicle, or an autonomous robot. Current day robotics demand the need for robust road detection in structured and unstructured roads, owing to emerging applications such as fire safety, automated delivery and  autonomous driving. These such robots need to be able to adapt to different scenarios and be capable in traversing in unstructured roads. Most autonomous systems which travel on road make high level decisions (direction to travel/obstacles to avoid) once the road is detected. Road detection may also be used in driver assistance systems, to prevent accidents and human errors.
In urban scenarios, detecting the road could be straightforward - as roads and elements on the road (lane markings, signs) have a specified structure. However, the same might be difficult in a rural or an unstructured environment owing to landscape roads. Such roads do not have definite structure, lane markings, surroundings or structure, and as a result it might not be wise to rely on traditional approaches \cite{sun2002road}.

\begin{figure}
	\centering
	\includegraphics[width=15cm]{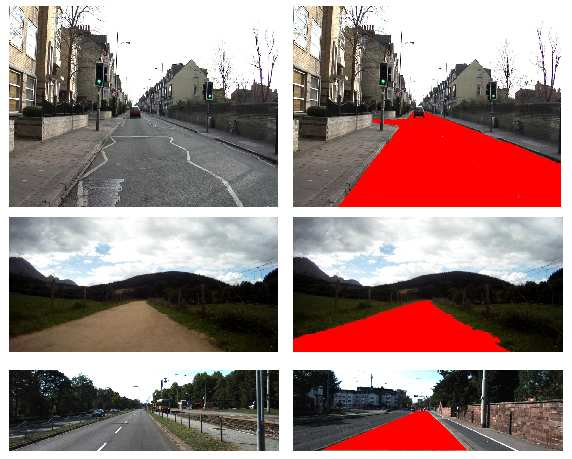}
	\caption{\textbf{The Task of road detection}. Left: Images from different scenarios (CamVid, Bilbao Raw and Kitti-Road) . Right: Overlayed ground truth}
	\label{gttask}
\end{figure}

The methodology on which road detection is done is highly dependent on the hardware present on the robotic or automobile system. The vision system could consist of a single camera, a stereo camera rig or a combination of multiple visual or multi-spectral sensors. Depending on the hardware, road detection is usually carried out by one of the two methods  - Monocular vision \cite{sotelo2004color,mohan2014deep,wang2013fast,xiao2015fast,kong2010general,choi2012road} based and multiple view based (stereo/lidar) \cite{xiao2015crf,vitor20132d,alvarez20103d,loose2010b}. In this paper, we focus on monocular vision. Traditional monocular vision based methods are performed pixel-wise, i.e, they observe only a pixel or its very local neighbourhood. In traditional approaches, the  features which are used for road detection are also hand-crafted(LBP/SIFT/HOG). They come with a disadvantage of being very specific and therefore are only able to discover specific and limited variations in an image. The predictions usually coming out of these methods are highly noisy, due to the ambiguity of the appearance of a single pixel. These noisy predictions are then traditionally passed on the graphical models - Conditional Random fields or Markov fields \cite{xiao2015crf}. CRFs and MRFs are models in which noisy predictions are taken into a pre-specified graphical structure, and then optimized based on some priors on neighbourhood interactions. This occassionally introduces oversmoothing when there is concept drift, and "cleans" up valid predictions. CRFs are computationally intensive as it performs iterations of updating neighbours of pixels based on current pixel values. Therefore it is slow, and is also able to create new errors due to unintentional smoothing of the predictions. 

Presently, road detection may also be carried out using Convolutional Neural Networks(CNN). Convolutional Neural Networks achieve high accuracy, as well as are fast owing to it's easy implementation on the GPU. However, CNNs have a  high footprint in terms of memory and computational requirement \cite{mohan2014deep}. In addition,  CNNs require large amounts of data to achieve high accuracy without overfitting \cite{RosCVPR16}. 

In order to alleviate the disadvantages of existing traditional methods, and to the high footprint of CNNs, we propose a new pipeline using features coming from pre-trained CNN kernels and a strong classifier consisting of a random forest of local experts. We use superpixels instead of pixels, machine learned feature maps instead of traditional hand-crafted features. We use a pre-trained convolutional network (VGG-16) to extract features. To classify these descriptors, we use a random forest framework, modifying the traditional threshold based split function, to introducing Support Vector Machines in the nodes of the Random Forest - which is done by repurposing the random forest of local experts for pedestrian detection in \cite{marin2013random} for road detection. Furthermore, in our framework, the option of a custom grouping of heterogenous features is available - in the scenario where we need to combine machine learned features with traditional hand-crafted features.

We also introduce a dataset of unstructured road images, owing to the relatively newer application of vision based unstructured road detection. We perform the hyperparameter optimization of our method on the well benchmarked structured scene datasets (KITTI-Road \cite{fritsch2013new} and CamVid \cite{brostow2009semantic}). We perform experiments under similar settings in our unstructured road datasets. We also show the advantage in terms of memory footprint and training time. Furthermore, we also show how our classifier generalizes easily, by performing experiments on lesser training images.
\footnote{The Dataset is available online at \url{www.prassanna-ravishankar.github.io/LandscapeDataset} and the code for the random forests is available at \url{https://github.com/prassanna-ravishankar/Slither} . The code for the entire pipeline shall be uploaded soon and will be linked the landing page for the LandscapeDataset}

The paper is organized as follows - In Section 2 , we introduce the related work for road detection and random forests. In section 3, we explain our pipeline. In Section 4, we explain the experiments undertaken and the results, followed by future work in section 5. After this, we include a short section on our experimental setting. Finally, we draw our conclusions in section 6

\section{Related Work}
\label{sec:related-work}
There exist different methods of road detection depending on the hardware present. In the past, a stereo camera rig has been used to detect roads \cite{vitor20132d, loose2010b, sappa2007road}. They usually work by first combining the different views coming from the two different cameras, and then assuming that the road would have a smooth disparity map. More intermediate descriptors may also be obtained from the two views (such as stixels \cite{badino2009stixel}), on which the prediction may be made. This may be extended for multiple cameras using multi view geometry \cite{tupin2002road}. Furthermore, data from multiple sensors may be fused to create a "heat map" of combined features. The prediction is then made on these combined features or these features are independently introduced into a contextual model which combines them \cite{xiao2015crf}. However, recently, road detection algorithms have been performed on just a monocular camera input. With a monocular camera, the challenges are greater - disparity is ambiguous, occlusion is not handled organically and no sources of improving on errors from other views. On the other hand, monocular vision based approaches are significantly faster \cite{xiao2015fast}, because they do not have a pre-processing step. Monocular vision based approaches also require less hardware resources ( just one visual range camera) and thereby are more affordable.

Monocular vision based road detection algorithms may be classified into a) Traditional approaches using hand-crafted features or b) Approaches using Convolutional Neural Networks (Machine learned features).The approach specified in a) usually apply specific hand-crafted algorithms for feature extraction and computation, such as SIFT, LBP or HOG, amongst others \cite{ladicky2010and,zhou2010road} or filter banks \cite{sun2002road}. Hand-crafted features are traditionally used as they obtain features which are robust to rotation and scale which may be used for a variety of applications. In a road detection pipeline,  pixel-wise or region-wise hand crafted features are passed on to a classifier to apply a road or a non-road label \cite{jin2014hand}. However, hand-crafted features are very specific, and do not have the ability to adapt their representations with data \cite{alexnet}. For example, the feature descriptors coming out of a road in sunlight, and a road in rain could be significantly different.

It may be seen that the accuracy of road detection may be improved by using a contextual model. Context becomes especially useful in different situations of weather, illumination or appearance. This may be introduced by stacking this with a conditional or a markov random field, which learns pairwise dependencies from data and smooths the prediction based on that. While random fields improve the results usually, they occassionally smooth out discontinuities in pixel predictions \cite{crfroaddetection}. Alternatively, instead of labeling each pixel, a grouping of pixels may be labeled. This grouping could be done by a) dividing the image into fixed size \textbf{patches} or b) computing \textbf{regions }based on pixel similarities.\textbf{ a) Patch based approaches} have been used before \cite{xiao2016monocular}, entire patches are looked upon for different kinds of road boundary shapes which were earlier stored in a dictionary. It is usually used to classify road edges, after which everything within this region is considered as road. However, there can be drawbacks to these methods in the scenario of unstructured roads where road boundaries are not fixed patterns, or in the case of occlusion where an object of an unknown (or not learned) shape lies on the road. \textbf{b) Superpixels} \cite{achanta2012slic} may also be used to segment the image, and then to classify each of these superpixels independently. Superpixels have been used for the purpose of road detection and it has been shown that especially in unstructured environments, superpixel based methods perform better than patch based methods \cite{superpixelsarebetter}. While a superpixel forces nearby similar pixels to take the same label, it often creates significant discontinuities and introduces artefacts in the edges between superpixels \cite{superpixellattices}. This is because superpixels are labelled with a single label which is then passed onto each pixel in a superpixel. In all of these methods, colour and colour statistics features may directly be used, or hand-crafted features, or machine learned features. The flexibility lies with the designer.

In contrast to Traditional hand-crafted features, Machine learned features may come from a deeply learned convolutional neural networks. A CNN tries to break down an image into a stack of kernels, each layer picking out spatial frequencies which repeat. The kernels in a deeply learned CNN are learned from a large dataset, thereby capturing spatial frequencies coming from thousands of images, from different sources and different backgrounds. Thereby, the CNN ultimately learns different kinds of edge and texture detectors in the early layers, and each successive layer, the more complex spatial kernels. Thereby a CNN eventually creates a stacked representation of the image data \cite{featurelearning}. CNNs maybe used as a complete pipeline (last layers of CNN acts as a classifier) to do end to end road detection, without extracting features from intermediate layers, or intermediate layer outputs could be extracted as features, with a classifier stacked onto it, to predict based on early features from a CNN. On one hand, CNNs are parameter-heavy and computationally-heavy and require relatively expensive hardware. On the other hand, features may also be extracted by using the convolutional kernels coming out of pre-trained networks \cite{hypercolumns}, which is relatively less taxing computationally (as no training is required).

It may be seen therefore, that no one kind of approach is best suitable for road detection. Each one carries it's advantages and disadvantages. In this paper, we stack different approaches, such that their drawbacks are not significantly error-prone. We first compute machine learned features, coming from the first layer of VGG-16, pre-trained on the ImageNet dataset \cite{imagenet}, to alleviate the problem of weak representation in hand-crafted features. Then, Superpixels are computed at multiple scales and the features are pooled into these superpixels, creating a superpixel lattice as in \cite{superpixellattices}, to alleviate the problem of lack of context. Then, each superpixel is classified and the predictions for all scales are pooled together, to alleviate the contour discontinuity in superpixels. The classifier we use is a random forest of local expert. Inspired by \cite{marin2013random}, our random forest of local experts is able to work with heterogeneous features, different CNNs or CNN layers and different modalities. At each node of the random forest, the split function takes into account different machine learned feature maps at random, and an SVM is locally applied on this. This method attempts to alleviate the drawbacks of hand-crafted features, by using machine learned features. It attempts to alleviate the drawbacks of machine learned features, by using only the early features. It further attempts to alleviate superpixel artefacts by using multiple scales of superpixels. Furthermore we use a cascaded ensemble of SVMs (strong classifiers), to strongly classify the "weak" feature maps coming out of the early layers.

In addition, we also release a dataset for unstructured road detection - "Bilbao Raw". There are only a few datasets of unstructured road that are available for public consumption\cite{xiao2016monocular}, and we would like to provide diversity to the unstructured road detection challenge.

\begin{table}[h]
	\centering
	\caption{\textbf{Dataset Details} - This table describes the published dataset with respect to the other datasets. It may be seen that our dataset "Bilbao Raw" is best suited for Landscape scenarios where fixed urban assumptions cannot be made}
	\label{my-label}
	\begin{tabular}{llllll}
		\toprule
		\textbf{Criteria}          & \textbf{Kitti-UM} & \textbf{Kitti-UMM} & \textbf{Kitti - UU} & \textbf{CamVid} & \textbf{Bilbao Raw} \\		
		\midrule
		
		\textit{Urban environment} & \checkmark                  &  \checkmark                  &  \checkmark                   & \checkmark               &   \checkmark             \\
		
		\textit{Lane Markings}     &  \checkmark                & \checkmark                    &                     & \checkmark         &                     \\
		
		\textit{Street Signs}               &   \checkmark           &  \checkmark          &  \checkmark                   &  \checkmark               &  \\
		
		\textit{Different Ambient Conditions}               & \checkmark  &   \checkmark                 &   \checkmark                  &  \checkmark               & 
		\\		

		\textit{Continuous Road Boundries}               & 		 &                    &   				   &  & \checkmark
		\\			
		
		\bottomrule                   
	\end{tabular}
\end{table}

\section{Proposed Method}
\label{sec:proposed-method}

\begin{figure}[h]
	\centering
	\includegraphics[width=16cm]{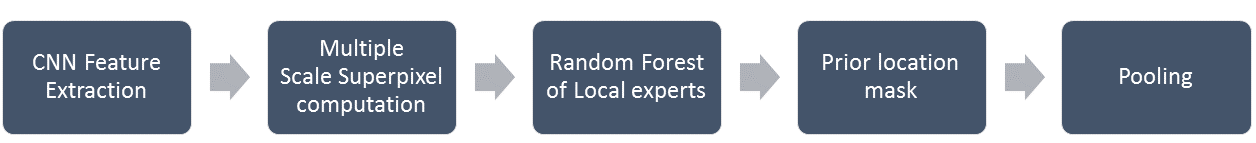}
	\caption{\textbf{Overview of Proposed Pipeline}}
	\label{proposedmethod}
\end{figure}

A semantic segmentation pipeline traditionally has two processes - Feature extraction and classification. In addition to this, they might further include a contextual model (such as a conditional random field) - which we do not use. We build our algorithm on the same pipeline, modifying to individual blocks. The features we extract are machine learned superpixel features. We extract features at various superpixel scales. The classification model we use is a random forest. The random forest we use is adapted to work with machine learned features, randomly selecting sub-features and testing it with an SVM. In the end we apply a location prior to smooth out our false positives. An overview of the proposed pipeline is given in Figure \ref{proposedmethod}. The following sections give an in-depth explanation over the individual blocks.

\subsection{Feature Extraction}
In this section, we focus on the feature extraction process. Features may be extracted image-wise, region-wise or pixel-wise. In the following subsections, we comment on the feature extraction method chosen.

\subsubsection{Machine Learned features}
\begin{figure}[h]
	\centering
	\includegraphics[width=16cm]{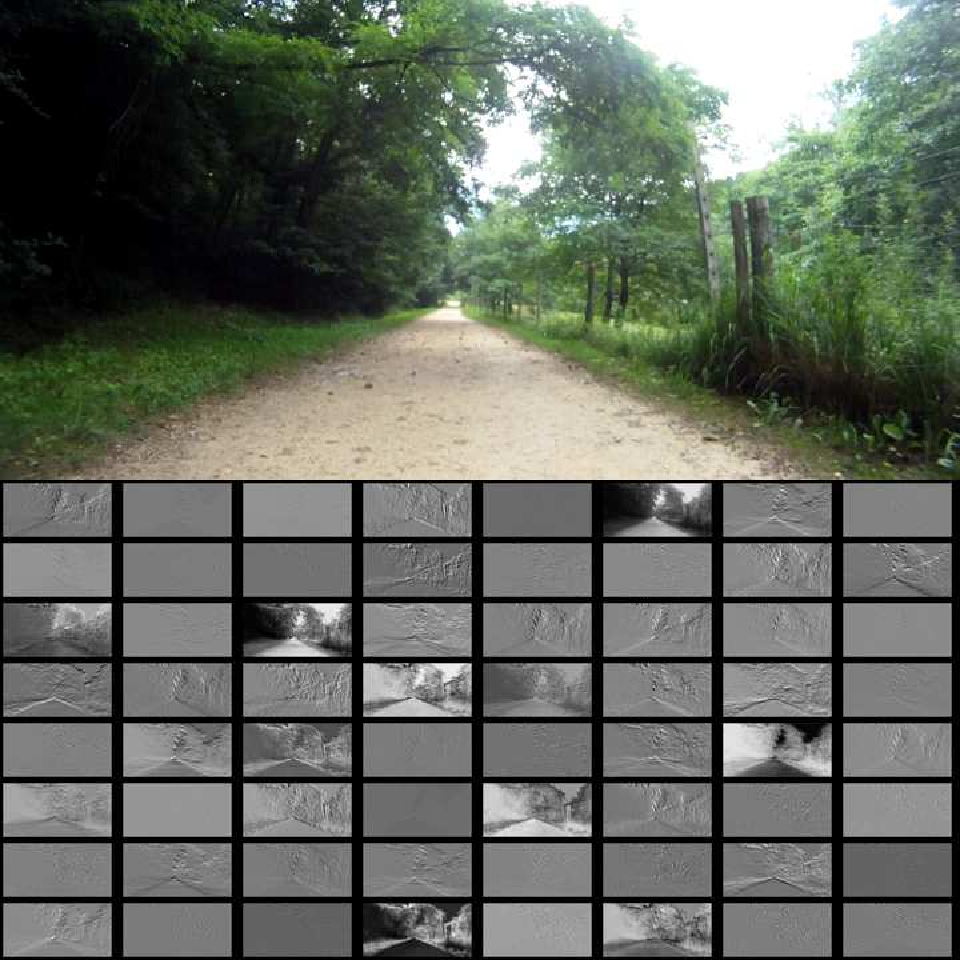}
	\caption{\textbf{Feature Maps from Bilbao-Raw} a) [TOP] A sample image from the unstructured dataset and b)[Bottom] Feature maps (64 in number arranged in 8x8) derived from convolutional layer 1 filters of VGG-16. Layer 1 of VGG-16 has 64 convolutional kernels, which result in 64 feature maps. It may be seen that some feature maps are more responsive to different orientations of texture/edges.}
	\label{featuremaps}
\end{figure}

Hand-crafted features refers to the task of computing features based on some fixed methodology. This fixed methodology is quite often manually designed. While such features work robustly in different scenarios of varying scale,rotation or texture, they are fundamentally incapable of adapting to a custom domain. On the other hand, Machine learned features are feature computation methods which may be adapted depending on the data. Therefore, often, they are domain specific. While machine learned features may be computed in many different ways, we limit ourselves to features learned from convolutional neural networks. The ideology behind this is that neural networks learn to break the image down into it's constituent spatial frequencies, which recur the most in the dataset that it sees. It thereby creates a stacked representation of the image \cite{featurelearning}. Therefore, theoretically, a CNN that learns from a large dataset of images (about 1 million images), will be able to learn commonly occuring low-level spatial frequencies. Therefore, the kernels learned in layer 1 of a deeply learned model would be commonly occuring edge detectors and basic texture detectors. As we do not have large datasets at our disposal, we use a pre-trained CNN for our application. We show some of the detected features from a pre-trained CNN  in Fig. \ref{featuremaps}. The process of choosing the convolutional layer and the network are further described in the experiments section.

\begin{figure}[h]
	\centering
	\includegraphics[width=5cm]{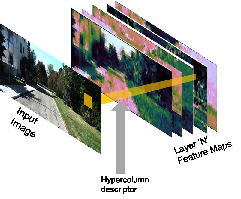}
	\caption{\textbf{Representation of a hypercolumn }of feature maps from a layer of a neural network}
	\label{hypercolumnIm}
\end{figure}

We use the kernels that a convolutional neural network learns to calculate the feature descriptors for an image. The image is passed through VGG-16 \cite{chatfield2014return} and the outputs are taken at the first convolutional layer. This output constitutes the feature maps for the given input. The feature maps are then resized to the original spatial dimensions of the image, and are treated as dense features. Therefore,  we achieve a 64 dimensional pixel-wise feature descriptor, which we call as a \textbf{hypercolumn}, inspired by \cite{hypercolumns} and \cite{cimpoi2015deep}. The hypercolumn representation is qualitatively explained in Fig. \ref{hypercolumnIm}.

\subsection{Superpixel computation}
The feature maps coming out of the convolutional layer 1 of the Neural network are not ready for use in the classification module yet. Due to the presence of stride(s) in the convolutional layer and due to the max-pooling layers, the feature maps obtained are of a smaller spatial resolution than the input image. This, when resized to the same spatial dimensions, introduces artifacts. Furthermore, predictions/classifications done on the pixel level do not consider the local neighbourhood of that pixel. 

To tackle this, we use superpixels to pool in the hypercolumns. We compute superpixels with the SLIC \cite{achanta2012slic} method. Superpixels are a grouping of pixels which are continuous in distance and colour. Grouping such pixels together, allows us to make one prediction per superpixel, and this prediction is passed onto the individual pixels. The statistics of a superpixel are captured depending on the feature vectors of the pixels it contains 
\begin{equation}
\xi
F_{\phi}(R_{j}) = T_{p_{i} \in R_{j}}[F(p_{i})]
\end{equation}
Where F is  a feature vector that corresponds to a pixel. $F_{\phi}$ is  a superpixel feature vector. $S_{j}$ is a superpixel region which consists of the pixels $p_{i}$. The function $T$ captures statistics over the corresponding feature vectors - We use mean($\mu$) and standard deviations($\sigma$) of the feature descriptors within a superpixel to capture the dominant feature vector and its variations, which are then concatenated.

However, this creates another problem of label confidences being discontinuous in superpixel edges - nearby superpixels could have very different predictions and could create unwanted edges on the resulting confidence maps. We resolve this problem by making predictions on multiple superpixel scales or superpixel lattices \cite{superpixellattices} and pooling them together. Different superpixel scales create different edge artifacts, which get smoothed out once pooled together.

\subsection{General Random Forest}
\label{general-random-forest}
We introduce the basic concepts and the notations of the Random Forest Ensemble \cite{breiman2001random} for background; the random forest we use is described in section \ref{local-expert-random-forest}. The random forest is a powerful classification model which stacks multiple weak classification models together. It works by initially introducing a weak learner to classify the given data. This divides the data into a positive branch and a negative branch (right) depending on the initial classification (left). A weak learner is applied to each of these branches, to further correct the false positives and the false negatives. This carries on till the maximum specified depth of the tree is obtained, or if the branching does not give us a gain(Gini criteria/Information gain) in accuracy. This consititutes the training of one tree. The same maybe done, with different initializations on more trees. A collection of such decision trees is called a random forest. To ensure brevity, we limit ourselves to the standard weak learner model and we refer to \cite{breiman2001random}for a more detailed description of the Random Forest classifier.
We proceed to mathematically represent the random forest. Given a tree of a forest, we can define a split node with index \textit{j}, which determines how the samples - $S_{j}$  received by the \textit{j}-th node are divided into it's branches. The splitting of samples $S_{j}$ is done by a split function $h(\vec{x},\theta_{j} ) \in   \{0,1\}$, where $\vec{x}$ is the feature vector and $\theta$ is the set of parameters defining the split function. Any classification model may be used as the split function, while the traditional random forest uses a decision stump.

\subsubsection{Node Functions}
To generalize the split function, we define the node parameters as $\theta_{j} = (\phi_{j}, \psi_{j}, \tau_{j})$ where $\phi$ is defined as a feature selection function that disregards any noisy features in $\vec{x}$. The feature selection function usually works in a random fashion, introducing the term "Random" in a Random Forest. The parameter $\psi$ defines a geometric transformation which transforms the data into a space where is separable. $\tau$ determines the threshold on basis of which that sample is classified. We ignore the index \textit{j} for simplicity.

\subsubsection{Training}
The training of a random forest takes a top-down approach.

\begin{enumerate}
    \setlength\itemsep{2em}
	\item For each node \textit{j} and the incoming training set $S_{j}$, we look for the best split function $h(\vec{x},\theta_{j} ) \in   \{0,1\}$ to split $S_{j}$ into $S_{j}^{L}$ and $S_{j}^{R}$, where:
	\begin{align*}
	    S_{j}^{L} = \{x \in S_{j} | h(\vec{x},\theta_{j}) = 0\} \\
        S_{j}^{R} = \{x \in S_{j} | h(\vec{x},\theta_{j}) = 1\}
	\end{align*}
	
	where $S_{j}^{L}$ and $S_{j}^{R}$ refer to the samples classified into the left and the right child branches respectively 
	
	\item The "goodness" of this partitioning is evaluated using some measure of purity, usually the information gain
    \begin{align*}
        I(\theta) = H(S_{j}) - \sum_{i \in (L,R)} \frac{S_{j}^{i}}{S_{j}}H(S_{j}^{i}) \\
        H(S) = -\sum_{c \in C}p(c)log(p(c))
    \end{align*}
        	where H(S) is the entropy
	
	\item 
	The parameters for the node j are then defined as $\theta_{j} = \argmax_{\theta \in \tau_{j}} I(\theta)$
		
	\item
	This process is repeated for every successive node till maximum depth has been achieved, or till a branching does not yield any gain.
\end{enumerate}

As a Random Forest is an ensemble model of \textit{N} decision trees,  each decision tree varies from the others owing to the injection of randomness during the training of each tree. This may be introduced via random training set sampling (bagging) and randomized node optimization (selecting different families of feature selection functions $\psi$ or thresholds $\tau$). Introducing randomness reduces overfitting and allows for the learned model to be generalized.

\subsection{Random Forest of Local Experts}
\label{local-expert-random-forest}
With respect to the Random Forest introduced in section \ref{general-random-forest}, instead of using a weak learner as the split model, we using a strong classification model(support vector machines) at the nodes - which we call as a \textbf{local expert}. While the general weak learner model of the Random Forest aims at improving branching based on an impurity measure, we extend this model to work by maximizing a purity measure and a maximum-margin optimizer which minimizes classification accuracy of node $S_{j}$. This is done by introducing $\psi$ to be based on the linear SVM learning algorithm. This extension is highly inspired by \cite{marin2013random}, and is further extended by us to be general purpose (and not limited to pedestrian detection with spatial blocks).

The feature descriptor input to the random forest is superpixel based hypercolumns. Since the hypercolumn is computed over a single layer in a convolutional neural network, each attribute of the feature vector is also the output of a specific kernel convolved with the input image. 

With respect to the model previously defined in Section \ref{general-random-forest}, we redefine the node functions for our Random Forest of local experts.

\begin{enumerate}
	\item \textbf{Feature Selection Functions} ($\phi$) \\
	The feature selection function is relatively unmodified. In case we have multiple statistics for a superpixel, we force the feature selection function to take all the specified statistics(mean and standard deviation) for a channel(output of the corresponding convolutional kernel). We select feature attributes in random and create a family of feature selection functions ($\{ \phi_{1}, \phi_{2} ........, \phi_{K} \}$). For each $\phi_{k}(\vec{x})$, we evaluate how good the feature selection function is.
	
	\item \textbf{Transformation Functions}($\psi$) \\
	For each feature selection function, we learn a linear transformation $\psi_{k}$, which transforms the received input $\phi_{k}(\vec{x})$ into a space where it is more linearly separable. To do this, we train a linear SVM over the transformed samples $S_{j}^{\phi_{k}}$. The function thus learned contains weights with which each feature attribute is scaled with and a bias. Therefore, $\psi_{k}(\vec{x}) = \textbf{b} + \vec{w}.\vec{x}$ which is the classification score of the linearSVM predictor.
	
	\item \textbf{Thresholding}($\tau$) and \textbf{purity measure} \\
	We then find a threshold ($\tau_{k}$) that maximizes the purity of the obtained partition. Therefore 
    \begin{equation}
	\begin{split}
	S_{j}^L = \{\vec{x} \in S_{j} : \psi_{j}(\phi_{j}(\vec{x})) \leq \tau_{k} \} \\ 
    S_{j}^R = \{\vec{x} \in S_{j} : \psi_{j}(\phi_{j}(\vec{x})) > \tau_{k} \} 
	\end{split}
    \end{equation}
	
	Then we let $P_{k} = I(\phi_{k}, \psi{k},\tau_{k})$ be the purity function which determines information gain. The threshold that maximizes the information gain is fixed as the threshold for the corresponding feature selection.

\end{enumerate} 

At  node $j$, after evaluating all the feature selection functions, we select the $\phi_{k}$ which maximizes the $P_{k}$ for the samples $S_{j}$. The biggest difference between our proposed model and the standard weak learner model is that we introduce a local expert that determines the best separating hyperplane at every evaluation. In the standard approach, split functions are generated at random, and the best one is selected - this may not be the optimal maximum-margin separation. In comparison, the linear SVM in the split function always attempts to obtain a hyperplane which best splits the data.

\subsubsection{Memory utilization}
At each node $[1,K_{max}]$ convolutional kernels (corresponding to the convolutional layer of the CNN) may be selected, where $K_{max}$ is the maximum number of Kernels available from the feature maps. For kernel, two superpixel statistics are computed - Mean and standard deviation. For each node, an svm is trained, whose weights are equal to the number of kernels selected, in addition to a bias. Let the size of each node by $M_{node}$. Therefore, $ M_{node} = (2K + 1 ) * sizeof(float) $

 Let the number of decision levels of the tree be $l$ and the number of trees be $T$. Let the maximum number of nodes in a tree be given by $N_{node}$, and memory requirement of a tree be given by $M_{tree}$, and a forest by $M_{forest}$. Therefore, 
\begin{equation}
 \begin{aligned}
	N_{nodes} = 2^{l} \\
	M_{tree} = N_{nodes}*M_{node} \\
	M_{forest} = T * M_{tree} \\
	M_{forest} = T * 2^{l}*(2K+1) *sizeof(float) \\
	M_{forest}^{max} = T*2^{l}*(2K^{max}+1) * sizeof(float) 
 \end{aligned}
\end{equation}

 Substituting the values for a 10 decision level forest with 10 trees, using the VGG Network (64 layer 1 kernels), and using 4 byte float, we get $M_{forest}^{max} = 5.039MB$. In reality, a trained network of these parameters comes to 3MB. This is because some nodes choose to select fewer than the maximum number of kernels, and some branches terminate before the maximum depth of the tree is reached.

\subsection{Pooling and Prior Location Mask}
We notice that our classifier introduces false positives. In order to reduce them, we apply a prior location mask. The location mask is learned from the training images in the image, and it stores a pixel wise prior probability based on location. We predict each superpixel  independently using the random forest classifier. However, we train and predict using different classifier models for different superpixel scale. This allows us to remove the edge artifacts on the predictions. We then pool the predictions for each scale, using a mean operation.

\section{Experiments and Results}
We present the results of our road detection system on three datasets. We apply our road detection on two problems - a) To detect and segment the road in structured road datasets (in urban, semi-urban scenarios) and b) To detect and segment the road in unstructured datasets (in non-urban scenarios where the road may or may not be made of asphalt). We use a) to do the hyperparameter optimization.

\subsection{Experimental Set-up}
The algorithm was implemented in python under Ubuntu 14.04. The Random Forest framework is in C++ with a python wrapper. Experiments were tested on a PC with 64GB RAM and Intel Core i7-5960X @ 3.00Ghz with 4 cores, hyper-threaded to 8 cores. We do not use a GPU. We implement the feature extraction and superpixel pooling algorithm in python and stack it with the Random Forest Framework, which is in C++ with a wrapper in python. Our algorithm benefits from a CPU with many cores, as the random forest method is highly paralleliz-able - evaluating multiple feature selections for each node in parallel, or training multiple trees in parallel. The code is published online at http://atemysemicolon.github.io/LandscapeDataset/. 

\subsection{Dataset details}
We evaluate our road detection method on two types of datasets - a) Structured (urban) road environments and b) unstructured (landscape) environments.  All our datasets only two labels(Road, Non-Road) in their ground truth images. If there are more, we map them to these two labels.

\textbf{Structured Road Datasets}: To test our algorithm in a structured environment, we evaluate against two datasets - Kitti \cite{fritsch2013new}, and CamVid \cite{brostow2008segmentation,brostow2009semantic}. We use the training set of the Kitti road dataset, and further split the dataset into a training set and test set, each containing ~150 images. Each image is approximately 1242x375 pixels, however varies slightly with the sequence selected (UM - Urban Marked, UMM - Urban Multiple Marked and UU-Urban Unmarked). For structured environments, we also test against the CamVid dataset. The CamVid dataset consists of 570 images divided into 4 sequences. We use 3 of these sequences in our training set and one unseen sequence for our test set. Each image in CamVid is 960x720 pixels. We also use the Kitti dataset to discover the optimal parameters for our method, and apply this against the other datasets. 

\begin{figure}[t]
	\centering
	\includegraphics[width=15cm]{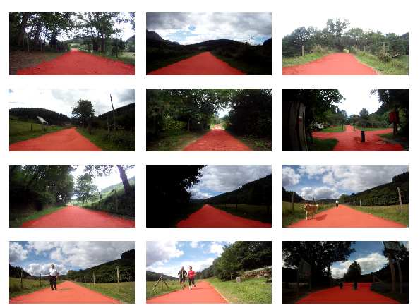}
	\caption{Some images from the bilbao-raw dataset. The road ground truth has been overlayed in red colour; The road is not red.}
	\label{unstructureddataset}
\end{figure}

\textbf{Unstructured Road Dataset}: Structured Road scenes are useful for evaluating the performance of an algorithm in urban-scenes. This is highly useful for the application of autonomous and assisted driving. However, an algorithm or a system trained on structured data might not perform so well in unstructured scenarios. Unstructured scenarios are common for robots or off-road driving, where roads are predominantly non-asphalt. In addition, urban roads often have lane markings which can be used for automated assistance, but landscape roads do not. Furthermore, there are different constraints for driving/navigation in unstructured scenarios - Sidewalks may not be present, a goat path may intersect with an asphalt road, there could be water or small obstacles on the path. To evaluate against unstructured environments, we introduce a new dataset - \textbf{Bilbao Raw}. At the time of writing, there is only one other unstructured dataset available - Freiburg Raw \cite{deepscene}. We nevertheless introduce another unstructured dataset to further community efforts. Bilbao Raw is a sequence of images in an unstructured, non-asphalt road environment, obtained in the country side of Basque Country. There is one sequence of 1000 images in the dataset, varying through slightly different ambient conditions - shadow, indirect sunlight and direct sunlight. There are also deformations and obstacles on the road. Each image in the dataset is 720x360 pixels.  We also allow for more images to be labelled by the community. Some images from this dataset are shown in figure \ref{unstructureddataset}. Our dataset is published at : \textbf{https://prassanna-ravishankar.github.io/LandscapeDataset/}

\subsection{Results}

\begin{figure}[p]
	\centering
	\includegraphics[width=15cm]{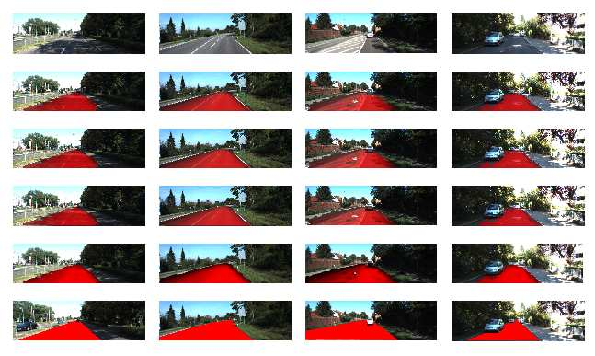}
	\caption{\textbf{Qualitative Results on Kitti-Road}. From Top to Bottom, row-wise (each column is a a different test image) - 1) Input Monocular Image 2) Prediction at 400 superpixels 3) Prediction at 800 superpixels. 4) Prediction at 1200 superpixels 5) Pooled prediction 6) Ground Truth. The prediction map is overlayed onto the input image, where the image intensity represents probability}
	\label{kittiresults}
\end{figure}

We compare our method to other methods benchmarked on the kitti dataset and present it in Table \ref{allresults} and in figure \ref{kittiresults}. It may be noticed that our method is better in performance in terms of pixel accuracy than traditional hand-crafted methods. At the same time, we are computationally less intensive and require less memory(~2mb), than Convolutional Neural Network based methods(VGG ~500mb \cite{vggnet}, SegNet\cite{segnet} ~200mb). The pipeline currently spends 3 seconds for feature extraction per image and superpixel pooling, which is the most time consuming step, the random forest predicts in 10ms. The average time taken to train (10 decision levels, 50 candidates) a tree using parallel cores in 0.7 seconds, for a dataset. This leads to the intuition that in order to save time, reducing the training set could suffice. A larger tree does not significantly increase training time when averaged over the number of images. However, as the tree branches are binary, each successive decision level theoretically takes twice the time to train as compared to it's parent. 

\begin{table}[p]
	\centering
	\caption{\textbf{Results on Kitti }(h-n) compared with Traditional Methods (d-g) and CNN based methods (a-c). (h-j) are experiments with lesser training images, being a variation of experiment l. The Maximum F score metric over different partitions of the dataset - Urban Marked(UU), Urban Multiple Marked (UMM) and Urban Unmarked(UU). We also show if the method was implemented on the GPU and CPU, and their respective runtime speeds.}
	\small
	\label{allresults}
	\begin{tabular}{llllll}
		\toprule
		\multicolumn{1}{l}{\textbf{Method}}                                                & \multicolumn{1}{l}{\textbf{UM}} & \multicolumn{1}{l}{\textbf{UMM}} & \multicolumn{1}{l}{\textbf{UU}}  &\multicolumn{1}{l}{\textbf{Speed}}  & \multicolumn{1}{l}{\textbf{Device}}\\ 
		\midrule
		a) Deep Deconvolutional Networks for Scene \cite{mohan2014deep} Parsing                                   & 93.65                            & 94.17                             & 91.76                            &2s  &GPU\\
		b) Multipurpose Deep Decoder Deconvolution  Network                                   & 93.99                            & 96.15                             & 93.69                            & 0.17s &GPU\\
		c) Neural Network plus Plane \cite{nnp}                                                         & 90.50                            & 91.34                             & 85.55                            & 5s  &GPU\\
		d) Superpixels and conditional random field with global shape prior  \cite{gheorghe2015semantic}                & 83.73                            & 87.96                             & 80.78                            & 2s  &CPU\\
		e) Graph Based Road Estimation using Sparse 3D Points from Velodyne \cite{shinzato2014road}                  & 85.43                            & 88.19                             & 84.14                            &60ms &CPU\\
		f) Histogram-Based Joint Boosting Classifier \cite{vitor2014comprehensive}                                        & 83.68                            & 88.73                             & 74.19                           &2.5min &CPU\\
		g) Structured random forest \cite{srf}                                                         & 76.43                            & 90.77                             & 76.07                            &0.2s &CPU\\ 
		\multicolumn{1}{l}{h)\textbf{ Machine Learned Descriptors - 10 Training Images}}                           & \multicolumn{1}{l}{87.44}       & \multicolumn{1}{l}{90.05}        & \multicolumn{1}{l}{78.09}       &3s &CPU\\
		
		\multicolumn{1}{l}{i)\textbf{ Machine Learned Descriptors - 50 Training Images}}                           & \multicolumn{1}{l}{88.98}       & \multicolumn{1}{l}{91.83}        & \multicolumn{1}{l}{81.49}       &3s &CPU\\
		
		\multicolumn{1}{l}{j)\textbf{ Machine Learned Descriptors - 100 Training Images}}                           & \multicolumn{1}{l}{88.69}       & \multicolumn{1}{l}{92.15}        & \multicolumn{1}{l}{81.28}       &3s &CPU\\

		\multicolumn{1}{l}{k)\textbf{ Machine Learned Descriptors + LBP}}                           & \multicolumn{1}{l}{90.65}       & \multicolumn{1}{l}{90.17}        & \multicolumn{1}{l}{83.40}       &4s &CPU\\ 
		\multicolumn{1}{l}{l) \textbf{Machine Learned Descriptors (VGG trained on ImageNet)}}       & \multicolumn{1}{l}{90.63}       & \multicolumn{1}{l}{92.79}        & \multicolumn{1}{l}{82.41}       &3s &CPU\\ 
		\multicolumn{1}{l}{m) \textbf{Machine Learned Descriptors (VGG trained on Places Dataset)}} & \multicolumn{1}{l}{89.14}       & \multicolumn{1}{l}{91.49}        & \multicolumn{1}{l}{81.39}       &3s &CPU\\ 
		\multicolumn{1}{l}{n) \textbf{Machine Learned Descriptors (SegNet)}}                        & \multicolumn{1}{l}{88.66}       & \multicolumn{1}{l}{91.51}        & \multicolumn{1}{l}{80.45}       &3s &CPU\\ 
		\bottomrule
	\end{tabular}
\end{table}

Furthermore, Our random forest of local experts is designed to allow for the user to decide how the features are grouped. We allow for a group of feature attributes to be selected together, and we allow for different feature attributes to be sampled at different frequencies. The effect of this is that we can fine-tune the training of the random forest. It also allows for different modalities or additional features to be added later, and combined with the current system, without re-training the entire system. We further test our algorithm trained with lesser training data. The results of this are shown in Table \ref{allresults} along with the other results. We show that we can reduce the training time without a signficant loss in accuracy. With the same setting of our parameters as obtained over the Kitti-Road dataset, we perform an experiment on the CamVid dataset. The results are in Table \ref{camvidresults}.

\begin{table}[p]
	\centering
	\caption{\textbf{Pixel-wise measures on the CamVid Dataset}. Interestingly, lowering the number of images does not reduce accuracy.}
	\label{camvidresults}
	\begin{tabular}{llllll}
		\toprule
		\textbf{Number of Images} & \textbf{} & \textbf{Accuracy} & \textbf{MaxF} & \textbf{Recall} & \textbf{Precision} \\
		\midrule
		10               &  & 92.82    & 94.99 & 95.99  & 94        \\
		50               &  & 92.5     & 94.69 & 94.5   & 94.89     \\
		100              &  & 92.52    & 94.82 & 96.05  & 93.62     \\
		350              &  & 92.1         & 94.45      & 94.09       & 94.8         \\
		\bottomrule
	\end{tabular}
\end{table}

\begin{figure}[h]
	\centering
	\includegraphics[width=15cm]{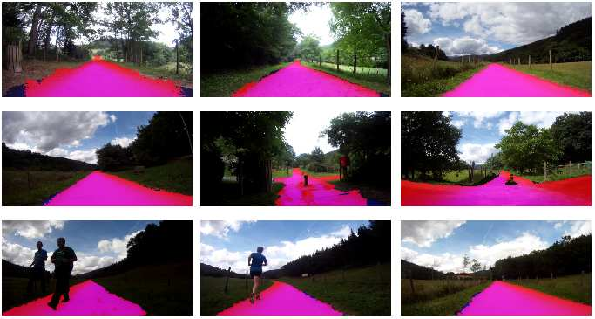}
	\caption{\textbf{Predictions on Bilbao raw} - predictions(blue) and Ground-Truth(red) overlayed on the original image. Regions where predictions match the ground truth are more pink in hue, false negatives are more red in hue and false positives are more blue in hue}
	\label{bilbaoraw}
\end{figure}

Selecting the same parameters as selected for Kitti, we train our random forest on unstructured data. The results of this are given in Table \ref{unstructuredresults}. We also show results on our method trained with significantly fewer images. We notice that we can make a good trade off by using less images to train. Furthermore, we show some qualitative results in Figure \ref{bilbaoraw}.

\begin{table}[h]
	\centering
	\caption{\textbf{Pixel-wise measures on Bilbao-Raw dataset}}
	\label{unstructuredresults}
	\begin{tabular}{llllll}
		\toprule
		\textbf{Number of Images} & \textbf{} & \textbf{Accuracy} & \textbf{MaxF} & \textbf{Recall} & \textbf{Precision} \\
		\midrule
		10                        &           & 96.38             & 97.85         & 97.8            & 97.89              \\
		50                        &           & 98.3              & 98.98         & 98.75           & 99.23              \\
		100                       &           & 98.3              & 98.99         & 98.86           & 99.12   \\  
		350						  &			  &	 98.33 	& 99.00			& 98.932  		& 99.18		\\        
		\bottomrule
	\end{tabular}
\end{table}

\subsection{Ablation Studies}
Multiple parameters of the individual blocks in our pipeline need to be determined. We first start by fixing parameters for the feature extraction module, which also includes setting the parameters for the superpixel algorithm. This is followed by a hyperparemeter search done on the Random Forest to discover it's optimal parameters.

\subsubsection{Design Choices for Feature Extraction}
\begin{figure}[h]
	\centering
	\includegraphics[width=10cm]{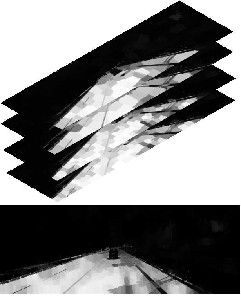}
	\caption{Multiple scales of superpixel pooled together to remove the superpixel contour effect. It may be seen that sharp discontinuities in the maps above have been "fuzzed" out}
	\label{superpixelpooled}
\end{figure}

Our method works on superpixels on a single scale, however, this introduces artifacts at the superpixel boundaries. In order to alleviate this problem, we make predictions on multiple scales and pool them together. By this, our superpixel boundary artifacts get "fuzzed away". Qualititative Results for individual superpixel scales may be seen on Figure \ref{superpixelpooled}. We fix the number of scales to be 3, and the use 400,800 and 1200 superpixels for the different scales.

We first decide which Convolutional Neural Network to use. This design choice is essential, as different combinations of networks trained with different training data, as each of them have different weights for their convolutional kernels. Since we are using these weighted convolutional kernels as filter banks, we effectively have a different family of filter banks to test each time. We select VGG \cite{vggnet}, as it gives us the best performance out of the networks we tested. The experiments to discover this are presented in Table \ref{ArchTable}. Furthermore, we also need to discover which convolutional layer of this network performs better for our application - the results of this experiment are in Table \ref{LayerTable}.We also use conv 1 of VGG as it is the best performing layer.

\begin{table}[h]
	\centering
	\caption{\textbf{Evaluation of Different CNN Models and initialization}. For all experiments, the random forest parameters were kept constant, and superpixel scales were 400,800 and 1200 }
	\small
	\label{ArchTable}
	\begin{tabular}{llllll}
		\toprule
		\textbf{Network}    & \textbf{Maximum F1 Score} & \textbf{Average Precision } & \textbf{Recall} & \textbf{False Positive Rate} & \textbf{False Negative Rate} \\ 
		\midrule
		\textbf{VGG \cite{vggnet}}        & 90.63         & 92.23                         & 91.34           & 4.71                         & 8.66                         \\
		\textbf{Places-VGG \cite{placesnet}} & 89.14         & 91.78                         & 90.20           & 5.61                         & 9.80                         \\
		\textbf{SegNet \cite{segnet}}     & 88.66         & 91.48                          & 88.5555         & 5.16                         & 11.45                        \\
		\textbf{OverFeat \cite{overfeat}}   & 87.74         & 90.81                         & 88.16           & 5.88                         & 11.84     \\                  
		\bottomrule
	\end{tabular}
\end{table}

\begin{table}[h]
	\centering
	\caption{\textbf{Evaluation of various convolutional layers of VGG}. All metrics are in percentage}
	\small
	\label{LayerTable}
	\begin{tabular}{llllll}
		\toprule	
		\multicolumn{1}{l}{\textbf{Layer}} & \textbf{MaxF} & \textbf{AvgPrec} & \textbf{Recll} & \textbf{FPR} & \textbf{FNR} \\ 
		\midrule
		1                                   & 92.08         & 93.70                       & 93.50            & 10.29            & 6.49             \\
		2                                   & 90.75         & 93.27                     & 92.58            & 12.3             & 7.41             \\
		3                                   & 89.14         & 93.00                      & 91.22            & 14.49            & 8.77             \\
		4                                   & 89.76         & 93.23                      & 90.49            & 12.13            & 9.50          \\
		\bottomrule  
	\end{tabular}
\end{table}

\subsubsection{Design Choices for Random Forest}

After selecting the Network, and it's corresponding layer, we run a hyperoptimization experiment to vary the depth, number of candidate features and the maximum number of decision levels of trees in the random forest. The result of the following is given in Table \ref{allresults}. We also fix the C Parameter of the SVM as C=0.5, after doing a grid-search experiment and varying C. 

\begin{table}[h]
	\centering
	\caption{\textbf{Hyperoptimization of Random Forest Parameters}. All Metrics are in percentage, truncated to two decimal places}
	\small
	\label{rfparams}
	\begin{tabular}{lllllll}
		\toprule
		\textbf{Depth} & \textbf{Candidates} & \textbf{Trees} & \textbf{Accuracy-UM} & \textbf{Accuracy-UU} & \textbf{Accuracy-UMM} & \textbf{Average Accuracy} \\
		\midrule
		5              & 10                  & 10             & 88.23        & 80.29        & 89.66         & 86.06             \\
		5              & 10                  & 100            & 87.96        & 80.46        & 89.62          & 86.01              \\
		5              & 100                 & 10             & 88.33        & 80.53        & 90.48          & 86.45             \\
		5              & 100                 & 100            & 88.41        & 80.35        & 90.76         & 86.51             \\
		5              & 500                 & 10             & 88.06        & 80.86        & 90.92         & 86.61             \\
		5              & 500                 & 100            & 88.30         & 80.40        & 90.76         & 86.49             \\
		10             & 10                  & 10             & 88.95         & 81.56        & 92.15         & 87.55             \\
		10             & 10                  & 100            & 88.67        & 81.37        & 91.45          & 87.16             \\
		10             & 100                 & 10             & 88.92        & 81.00        & 92.21         & 87.38             \\
		10             & 100                 & 100            & 88.52        & 81.28        & 92.16         & 87.32             \\
		10             & 500                 & 10             & 88.93        & 80.90        & 92.01         & 87.28               \\
		10             & 500                 & 100            & 88.67        & 81.85        & 91.99         & 87.50             \\
		15             & 10                  & 10             & 89.01        & 81.49        & 91.97         & 87.49             \\
		15             & 10                  & 100            & 88.40        & 81.70        & 92.02         & 87.38             \\
		15             & 100                 & 10             & 88.80        & 81.31        & 92.25         & 87.45             \\
		15             & 100                 & 100            & 88.70        & 81.17        & 92.14         & 87.34             \\
		15             & 500                 & 10             & 88.72        & 81.36        & 92.03         & 87.37   \\
		\bottomrule         
	\end{tabular}
\end{table}


\section{Discussion}
\begin{figure}[h]
	\centering
	\includegraphics[width=15cm]{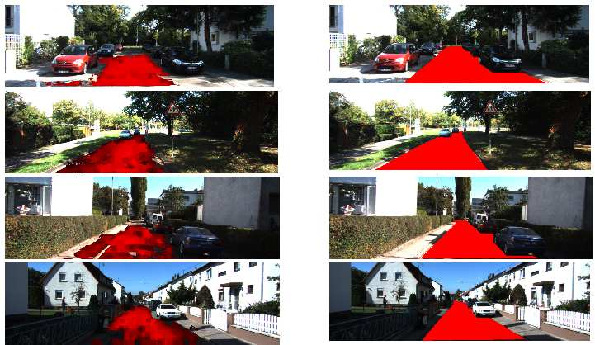}
	\caption{\textbf{Difficult cases for our proposed method}. Left- Predicted Image, Right - Ground-Truth Image. From top to bottom, in the first row the negative effect of high exposure may be seen. In the second, shadows affecting the road accuracy might be seen. In the third and the fourth row, pixels sidewalks are predicted and vice versa respectively due to shadows.}
	\label{errorim}
\end{figure}

There are some images which present difficult scenarios for our algorithm to predict. Some such cases are shown in Figure \ref{errorim}. This happens when the texture of the road is difficult to obtain or highly similar to other objects in the scene. One of these cases is when the sidewalk is similar in appearance to the road, in which case a part of the sidewalk is predicted as the road. Another scenario is when there are shadows on the road - in which case texture from the darker parts of the road are difficult to obtain and are predicted wrong. Also, the oversegmentation algorithm fails to follow the contours of the road, in which case a superpixel might contain a part of the sidewalk and result in wrong predictions. Surprisingly, most of our difficult cases are in structured datasets. In unstructured datasets, our algorithm performs exceptionally well, even after training it with just 10 images.
We consider about improving these results by either introducing more modalities (such as lidar or stereo), or by introducing a pixel level update in addition to the superpixel update.

\section{Conclusions}

In this paper, we show a simple way to combine Machine Learned features into a traditional pipeline to do Road Segmentation. In particular, we effectively use a pre-trained CNN with a strong classifier (Random forest of local experts). We incorporate the random forest in such a way that it randomly chooses the feature maps from the CNN. We predict for superpixels as the individual entity to gain local continuity and smoothness. Through our experiments, we show how our method lies very close to CNNs in terms of accuracy. However, we show that the proposed method has the ability to learn with very few training images, and is a lesser computational burden as compared to CNNs. Furthermore, we have publicly introduced a unstructured road dataset-Bilbao Raw. This allows us to test the feasibility of our method on robotic applications. We show that we achieve very good results with model parameters set with the assistance structured dataset. Finally, we also mention the possible avenues through which this work may be extended.

\section*{Acknowledgments}
This work was made possible due to the pre-doctoral PIF grant with reference number - RD 1393/2007.  We also thank the Computer Vision Center, Barcelona for their resources. 

\bibliographystyle{unsrt}  
\bibliography{references}

\end{document}